\documentclass{article}

\usepackage[nonatbib]{neurips_data_2023}

\usepackage[utf8]{inputenc} 
\usepackage[T1]{fontenc}    
\usepackage{hyperref}       
\usepackage{url}            
\usepackage{booktabs}       
\usepackage{amsfonts}       
\usepackage{nicefrac}       
\usepackage{microtype}      
\usepackage{multirow}
\usepackage{graphicx}
\usepackage{float}
\usepackage{subfig}
\usepackage{overpic}
\usepackage{array} 
\usepackage{threeparttable}
\usepackage{amssymb}
\usepackage{mathtools}
\usepackage{cite}
\usepackage{booktabs}  
\usepackage{tabularx}
\usepackage{authblk}
\usepackage{tikz} 
\usepackage{listofitems} 
\usepackage{amsmath}

\title{A Versatile Multi-Agent Reinforcement Learning Benchmark for Inventory Management}

\author{
Xianliang Yang$^1$, Zhihao Liu$^2$, Wei Jiang$^3$, Chuheng Zhang$^1$, Li Zhao$^1$, Lei Song$^1$, Jiang Bian$^1$ \\
$^1$Microsoft Research Asia \\
$^2$Institute of automation, Chinese academy of science \\
$^3$University of Illinois Urbana-Champaign \\
\texttt{\{xianliang, jiang.bian\}@microsoft.com }
}

\begin{document}

\maketitle
\nolinenumbers

\begin{abstract}
Multi-agent reinforcement learning (MARL) models multiple agents that interact and learn within a shared environment. This paradigm is applicable to various industrial scenarios such as autonomous driving, quantitative trading, and inventory management. However, applying MARL to these real-world scenarios is impeded by many challenges such as scaling up, complex agent interactions, and non-stationary dynamics. To incentivize the research of MARL on these challenges, we develop MABIM (Multi-Agent Benchmark for Inventory Management) which is a multi-echelon, multi-commodity inventory management simulator that can generate versatile tasks with these different challenging properties. Based on MABIM, we evaluate the performance of classic operations research (OR) methods and popular MARL algorithms on these challenging tasks to highlight their weaknesses and potential.
\end{abstract}

\section{Introduction}
Reinforcement learning (RL) is a critical branch of machine learning that aims to make a sequence of optimal decisions to maximize rewards\cite{sutton2018reinforcement}. It demonstrates remarkable success in various game domains, surpassing human performance in games like Go\cite{silver2017mastering}, StarCraft\cite{vinyals2019grandmaster, vinyals2017starcraft, shao2018starcraft}, and DOTA2\cite{berner2019dota, ye2020towards}. Besides gaming, RL is widely used in diverse domains such as industrial production\cite{spielberg2017deep, nian2020review, adam2011experience}, energy control\cite{perera2021applications, tittaferrante2021multiadvisor}, autonomous driving\cite{kiran2021deep, zheng2013decision, gao2019decision}, quantitative trading\cite{kolm2020modern, charpentier2021reinforcement, hu2019deep}, and recommendation systems\cite{chen2019generative, tang2019reinforcement}. The RL research is closely tied to the availability of suitable environments, which is well-established in various RL fields, including Multi-Agent Particle Environment\cite{gupta2017cooperative} and CityLearn\cite{vazquez2019citylearn} for energy control, SUMO\cite{krajzewicz2010traffic} for autonomous driving, Qlib\cite{yang2020qlib} for finance, and RecoGym\cite{rohde2018recogym} for recommendation systems.

Multi-Agent Reinforcement Learning (MARL) is a sub-field of RL that focuses on studying the behavior of multiple agents coexisting and interacting with each other in a shared environment. Due to its ability to model complex interactions and adapt to dynamic situations, MARL can be applied to real-world production scenarios where diverse and complex decisions are made simultaneously\cite{bucsoniu2010multi}.

Despite its strong applicability, MARL encounters various challenges, including scaling up, agent interactions, and non-stationary contexts, among others~\cite{hernandez2019survey}. To address these challenges, significant research efforts have been dedicated to developing solutions\cite{yang2018mean,foerster2016learning,chu2020multi,pinto2017robust}. However, the absence of a comprehensive benchmark makes it difficult to effectively test various algorithms on diverse challenges, and prevents a thorough evaluation of algorithm performance across an extensive range of problems.
 
Inventory management is a crucial aspect of Operations Research (OR), encompassing a wide range of complex real-world scenarios that require decision-making. The variety of commodities, warehouse interactions, and demand fluctuations make it an ideal platform for MARL research. To leverage this, we develop a Multi-Agent Benchmark for Inventory Management (MABIM), simulating a multi-echelon multi-agent inventory environment based on the OpenAI Gym framework\cite{brockman2016openai}. MABIM captures the complexities inherent in MARL, enabling comprehensive comparisons among different MARL algorithms within a realistic production context. This benchmark promotes advancements in tackling the diverse challenges of inventory management while also providing a platform for assessing MARL algorithm performance across various tasks. The environment code is available at \href{https://github.com/VictorYXL/ReplenishmentEnv}{https://github.com/VictorYXL/ReplenishmentEnv}.

Our contributions could be summarized as follows:
\begin{itemize}
\item We develop MABIM, an efficient and flexible inventory management environment that utilizes real data from our partner in the retail and supports multi-echelon warehouses while handling massive commodities. MABIM serves as an open and effective benchmark for addressing inventory management challenges.
\item  Leveraging MABIM's flexibility, we simulate a wide range of MARL challenges, including scaling up, cooperation, competition, generalization, and robustness, further enhancing its applicability in various scenarios.
\item We conduct the performance evaluation of both classic OR algorithms and MARL algorithms on these challenging scenarios, providing an in-depth analysis of their individual strengths and weaknesses.
\end{itemize}

\section{Background}
In this section, we start by 
briefly introducing
the current challenges faced by MARL. 
Then, we present 
an
overview of the existing MARL benchmarks. 
Afterwards, we introduce the inventory management problem and provide a concise summary of existing inventory management benchmarks.

\paragraph{MARL challenges.}
MARL is a rapidly growing research field that concentrates on creating learning algorithms for multiple agents operating within a shared environment. Although it has experienced significant progress, several challenges continue to persist. These challenges involve scaling up to accommodate a large number of agents, effectively managing the intricate interactions between agents that encompass cooperation and competition, and addressing the non-stationary dynamics resulting from both environment and the interactive learning agents\cite{hernandez2019survey}.
\paragraph{MARL benchmarks.}
Current MARL benchmarks have certain limitations that can impede a comprehensive assessment and comparison of diverse algorithms. Table~\ref{MARL benchmark} compares common MARL benchmarks with ours, highlighting the maximum number of agents, dynamic contexts (or exogenous state variables, see e.g.,\cite{dietterich2018discovering,zhang2023generalizable}), and user customization flexibility. 
Previous benchmarks face challenges in accommodating scenarios with thousands of agents, incorporating dynamic contexts, and providing flexibility in a single framework. 
These limitations may impact the precision and applicability
for evaluating MARL algorithms, underscoring the importance of developing a more comprehensive and versatile benchmark in MARL research.


\begin{table}
    \captionsetup{font=small}
    \begin{minipage}{.55\textwidth}
	\caption{Comparison for different MARL benchmarks.}
	\label{MARL benchmark}
	\centering
    \resizebox{\textwidth}{!}{
    \begin{tabular}{lllcc}
        \toprule
         & \#Agents & Interaction & Contexts & Flexibility\\
        \midrule
        SMAC\cite{samvelyan2019starcraft} & 27 & Coop. & \\
        GRF\cite{kurach2020google} & 11 & Coop. & \\
        GoBigger\cite{zhang2023gobigger} & 24 & Coop.\&Comp. & & \checkmark \\
        MPE\cite{mordatch2018emergence} & 40 & Coop.\&Comp. & \\
        MA-MuJoCO\cite{peng2021facmac} & 6 & Coop. & \\
        NeuralMMO\cite{ha2022collective} & 1024 & Coop.\&Comp. & & \checkmark \\
        ManiSkill\cite{mu2021maniskill} & 2 & Coop. & & \checkmark \\
        MABIM(Ours) & 2000 & Coop.\&Comp. & \checkmark & \checkmark \\
        \bottomrule
    \end{tabular}
   }
   \end{minipage}
    \begin{minipage}{.44\textwidth}
	\caption{Comparison for existing inventory management benchmarks.}
	\label{inventory management benchmark}
	\centering
    \resizebox{\textwidth}{!}{
    \begin{tabular}{lccc}
        \toprule
         & Real data & Multi-echelon & Multi-SKU\\
        \midrule
        OR-Gym\cite{hubbs2020or} & & \checkmark & \\
        MARLIM\cite{Remi2022or} & \checkmark & & \checkmark \\ 
        IM Sim.\cite{sridhar2021simulation} & \checkmark & & \checkmark \\ 
        MABIM(Ours) & \checkmark & \checkmark & \checkmark\\
        \bottomrule
    \end{tabular}
   }
   \end{minipage}
\end{table}

\paragraph{Inventory management.}
Inventory management is a critical problem in the field of operations research, involving the efficient control of the Stock Keeping Units (SKUs) for storage, acquisition, and distribution to meet customer demands and optimize costs\cite{tayur2012quantitative}. The main objective of inventory management is to strike a balance between stock availability and storage costs while minimizing stockouts and overstocks\cite{toomey2000inventory}. One of the key challenges in this field is the implementation of effective replenishment strategies\cite{stadtler2014supply}. Effective inventory management can result in improved customer satisfaction, reduced operational costs, and enhanced overall business performance.

Classic OR algorithms are effective in specific inventory management scenarios. The base stock algorithm by Arrow et al. (1951)\cite{arrow1951optimal} maintains desired inventory levels by ordering when the inventory falls below a predetermined level, making it suitable for low order cost situations. Blinder et al. (1990)\cite{blinder1990inventory} propose the ($s, S$) algorithm to avoid frequent replenishment, triggering orders when stock falls below reorder point $s$ and replenishing to the maximum level $S$. The base stock (BS) and ($s, S$) algorithms are two popular OR algorithms and serve as baselines in our later experiments. See Appendix~\ref{OR algorithm} for more details.

Due to the generalization and applicability 
of reinforcement learning algorithms, 
they are increasingly being applied 
to inventory management problems. Examples include Deep Q-Network (DQN)\cite{dittrich2021deep}, QMIX\cite{rashid2020monotonic}, QTRAN\cite{son2019QTRAN}, IPPO and MAPPO\cite{zheng2020deep}, and CD-PPO\cite{ding2022multi}.
These RL algorithms demonstrate promising capabilities for addressing inventory management challenges and 
may provide
performance improvement and better adaptability.

\paragraph{Inventory management benchmarks.}
Despite the extensive research conducted on inventory management, there is a lack of comprehensive benchmarks in this domain. Table~\ref{inventory management benchmark} provides an overview of existing efforts in this area. 
The characteristics highlighted in the table not only align the benchmark more closely with real-world production scenarios but also lend themselves to be transformed into challenges for MARL algorithms effectively. 

\section{MARL formulation for inventory management}
In this section, we first introduce 
how the inventory management problem is modeled in our paper~(Section~\ref{inventory management model}),
including the structure of the multi-echelon system, the dynamic process for each time step, and the calculation of the evaluation metric (i.e., the profit).
Subsequently, we present 
the MARL formulation of this problem
(Section~\ref{MARL formulation}).

\subsection{Inventory management model}
\label{inventory management model}
\begin{figure}[h]
    \centering
    \includegraphics[width=\textwidth]{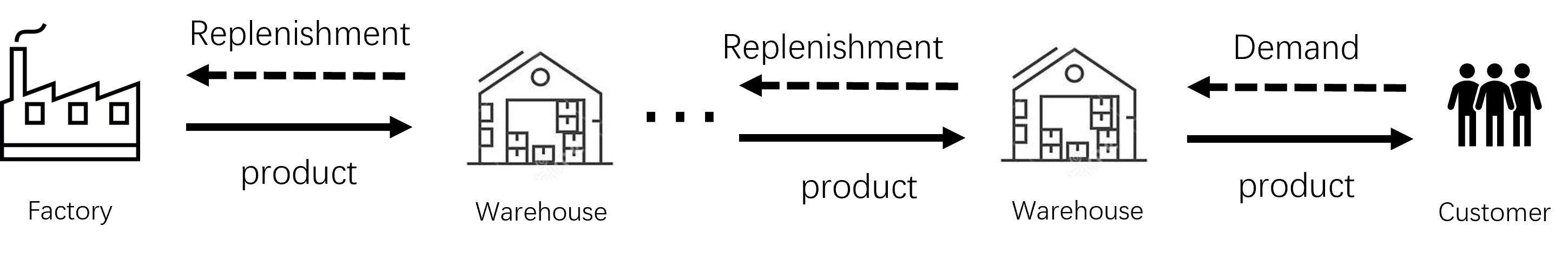}
    \caption{Multi-echelon inventory model.}
    \label{multi-echelon inventory management}
\end{figure}

\paragraph{Multi-echelon structure.}
We illustrate the multi-echelon model used in MABIM in Figure~\ref{multi-echelon inventory management}.
This modelling is motivated by the real-world process where products are produced by the factory and transmitted through echelons of warehouses sequentially until they reach the consumers. 
The goal is to optimize replenishment quantities for each restocking cycle (or time step), balancing inventory to avoid overstocking (causing increasing costs) as well as understocking (causing unmet demands).

\paragraph{Workflow.}
Each time step involves the agent making decisions regarding replenishment quantities for SKUs and subsequently transitioning the environment to a new state. Let $M\in\mathbb{Z}^+$ be the total warehouses, with the first one being closest to customers, and $N\in\mathbb{Z}^+$ the total SKUs. Given a variable $X\in\{D,S,L\ldots\}$, $X_{i,j}^t$ represents its value for the $j$-th SKU in the $i$-th echelon at step $t$, with $0\le i<M$ and $0\le j<N$. Given the above notations, the main progression of a step can be described as follows:
\begin{equation*}
\begin{aligned}
    \centering
    & D_{i+1,j}^{t+1} = R_{i,j}^{t} & \text{(Replenish)} \\
    & S_{i,j}^t = \min(D_{i,j}^t, I_{i,j}^t) & \text{(Sell)}\\
    & A_{i,j}^{t} = \sum_{k=0}^{t-1} \mathbb{I}(k+L_{i,j}^k==t) \cdot S_{i+1,j}^t & \text{(Arrive)}\\ 
    & \gamma_i^t = \min\left(\frac{W_i - \sum_{j} I_{i,j}^t}{\sum_{j} A_{i,j}^t}, 1\right), B_{i,j}^t = \lfloor A_{i,j}^t \cdot \gamma_i^t \rfloor & \text{(Receive)} \\
    & I_{i,j}^{t+1} = I_{i,j}^t - S_{i,j}^t + B_{i,j}^t & \text{(Update)}\\
\end{aligned}
\end{equation*}
Here, $D, R, S, I, A, B \in \mathbb{Z}^+$ and $\mathbb{I}(\text{condition})$ is an indicator function that returns 1 if the condition is true, and 0 otherwise.
\begin{itemize}
\item \textbf{Replenish:} Each warehouse requests a replenishment quantity $R$ for each SKU from the upstream source based on the policy, which becomes the demand of the upstream source.
The demand $D$ with $i=0$ comes from consumers, while other demands $D$ with $i>0$ come from replenishment orders of downstream warehouses.
\item \textbf{Sell:} Each warehouse sells the product to the downstream warehouse or consumers to meet their demands as much as possible.
Specifically, the sale quantity $S$ is set to be the demand $D$ capped by the current inventory level $I$. 
\item \textbf{Arrive:} Replenished SKUs arrive after lead time $L$ steps, which may be various. Thus, SKUs replenished at different steps may arrive simultaneously, so the arrival quantity $A$ is the sum of multiple previous replenishment quantities.
\item \textbf{Receive:} Limited warehouse capacity $W$ may prevent storing all arriving SKUs, causing overflow. The system allows custom acceptance strategies, and we present a built-in uniform receiving strategy in the above equation.
\item \textbf{Update:} The inventory for each SKU is updated after each step.
\end{itemize}

\paragraph{Profit.}
After each step, we calculate the profit generated by each SKU in each warehouse independently using the following formulation:
\begin{equation}  
\text{profit} =  
\underbrace{p \cdot S\strut}_{\text{Incoming}} - 
\underbrace{c \cdot S\strut}_{\text{Procurement cost}} - 
\underbrace{v \cdot (A-B)\strut}_{\text{Overflow cost}} - 
\underbrace{o \cdot \mathbb{I}(R > 0)\strut}_{\text{Order cost}} - 
\underbrace{h \cdot I\strut}_{\text{Holding cost}} - 
\underbrace{k \cdot (D - S)\strut}_{\text{Backlog cost}}
\end{equation}  
Here $p, c, v, o, h, k \in \mathbb{R}^+$ represent the unit selling price, procurement cost, overflow, order, holding, and backlog costs, respectively. We omit all subscripts denoting SKU, echelon, and step indices for better readability. Profit can serve as a useful metric for evaluating the effectiveness of a given strategy, and can also be used to design rewards in RL algorithm.

\subsection{MARL formulation}
\label{MARL formulation}
In our formulation, each SKU in each warehouse is modelled as an agent, which is responsible for making decisions regarding its replenishment amount in a warehouse. To ensure the environment is more scalable and adaptable to various scenarios, MABIM provides built-in functions for shaping actions, rewards, and observation states. Besides, interfaces are also available to customize them easily in order to accommodate specific needs and requirements.
\begin{itemize}
    \item The \textbf{observation state} for each agent is configurable, allowing for the inclusion of all current and past features of an agent. In addition to SKU features, warehouse or environment information can also be included, such as inventory occupancy and profitability.
    \item The \textbf{action} signifies the quantity of SKUs to be restocked. The system offers various built-in action-to-replenishment converters for enhanced generalization. In subsequent experiments, the average demand is used as a factor to establish the purchasing quantity.
    \item The \textbf{reward} is also configurable, with various built-in options provided. In the following experiments, we use the profit of each SKU in each warehouse as reward.
\end{itemize}
The details regarding observation state, reward, and action settings can be found in Appendix~\ref{MARL algorithm}.

\section{Core features of MABIM}
Building on the MARL formulation for inventory management, MABIM provides a high degree of flexibility to tackle different challenges frequently encountered in MARL research (Section \ref{tasks with configurable challenges}). To further enhance MARL research, MABIM also offers some other features such as efficiency, ease of use, and fidelity (Section \ref{features}).

\subsection{Tasks with configurable challenges}
\label{tasks with configurable challenges}

Compared to other inventory management environments, MABIM provides greater configuration flexibility, enabling the simulation of a wide range of challenges for MARL algorithms. This adaptability allows for customization tailored to specific use cases, enhancing the applicability of algorithms across various challenging tasks. Some typical challenges include:

\begin{itemize}
\item \textbf{Scaling up}: In the context of MARL, scaling up refers to the impact of numerous agents on training results and efficiency. While most recent MARL benchmarks do not support a large number of agents, our environment can support thousands of agents.

\item \textbf{Cooperation}: Cooperation between agents involves collaboration between warehouses at adjacent echelons. To satisfy consumer demands, deeper cooperation is needed in longer chains. Incentives, such as product profits or backlog penalties, ensure efficient product transfer.

\item \textbf{Competition}: Competition between agents arises when they vie for limited warehouse capacity. In our environment, reduced capacity or increased storage costs stimulate agents to compete for storage space.

\item \textbf{Non-stationary contexts}: Non-stationary contexts present a challenge for MARL, as they require the development of algorithms capable of learning in dynamic environments with fluctuating conditions. Examples include entirely new contexts to test the algorithm's \textbf{generalization} ability or noisy contexts to evaluate its \textbf{robustness}. These non-stationary factors may stem from external or internal sources. In the MABIM framework, the demand represents external context, and the features of SKUs serve as internal context.
\end{itemize}

\begin{figure}[h]
    \centering
    \includegraphics[scale=0.23]{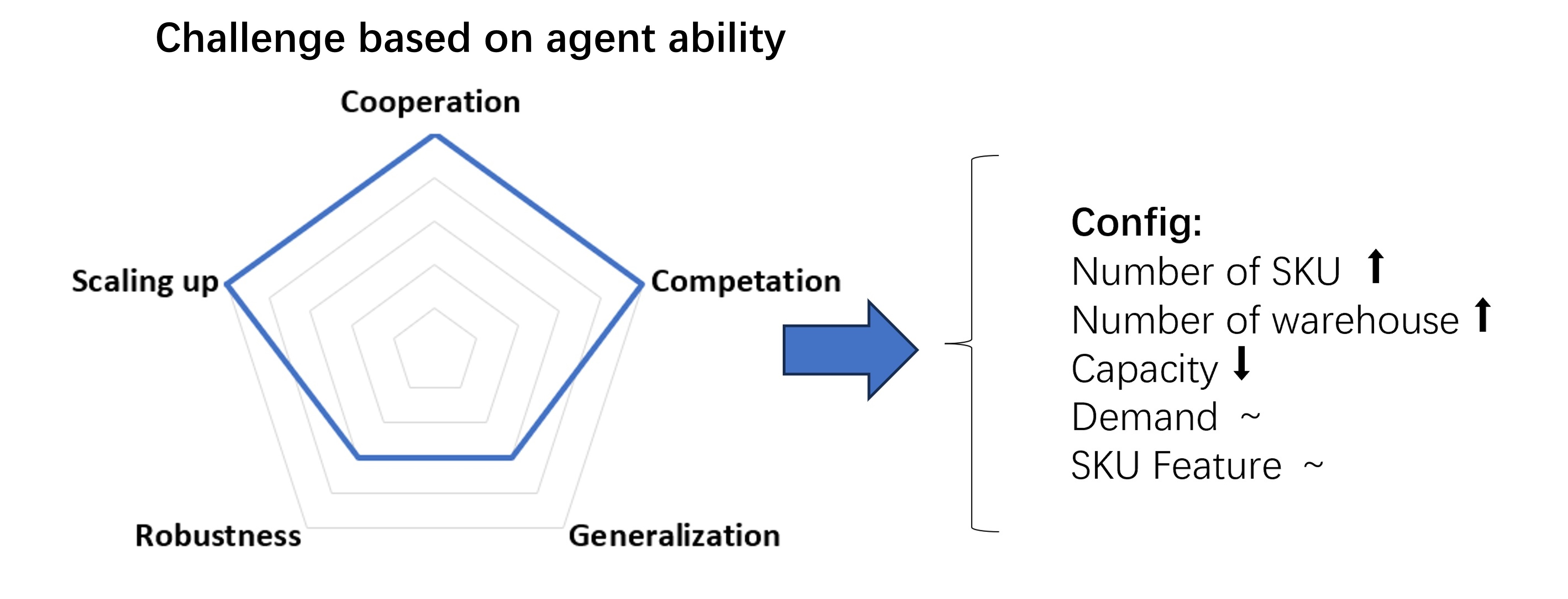}
    \caption{Combined challenge. $\uparrow$, $\downarrow$, and \textasciitilde{} signify increasing, decreasing, and keep steady.}
    \label{combined challenge}
\end{figure}

The challenges mentioned above can be managed by adjusting parameters, allowing for versatile combinations. Altering parameters such as the number of SKUs, warehouses, and warehouse capacity creates challenging environments to test competition, cooperation, and scaling up, as illustrated in Figure~\ref{combined challenge}.

MABIM contains a total of 51 built-in tasks with various challenges. For more details, refer to Appendix~\ref{task detail design}.

\subsection{Other Features}
\label{features}
In addition to processing various challenges with configurable difficulties, MABIM possesses several noteworthy features worth mentioning.

\paragraph{Efficient implementation.}
MABIM efficiently stores all SKU features and performs operations such as SKU initialization, purchasing, and selling in matrices. This approach ensures system efficiency and resource conservation.

\paragraph{Easy of use.}
MABIM utilizes a unified Gym\cite{brockman2016openai} interface and offers wrappers for common OR and RL algorithms. This consistency simplifies integration with other MARL frameworks and reduces the learning curve for researchers. Moreover, a visualization tool for analyzing SKU and warehouse states is provided. See Appendix~\ref{visulization} for more details.

\paragraph{High fidelity.}
MABIM aims to simulate real-world inventory management across various aspects to ensure more applicable solutions for actual production scenarios. Key features include the use of over 2000 \textbf{real demand data of SKUs}, running rule-based algorithms for \textbf{warmup} to avoid cold starts, providing different \textbf{overflow strategies} for processing overflow SKUs rather than discarding them directly, and incorporating an \textbf{acceptance strategy} that allocates storage space based on volume ratios to ensure fairness.

\section{Experiment}
\subsection{Experiment Settings}
During experiments, we compare the performance of OR and MARL algorithms. OR algorithms include base stock (BS) and ($s, S$), and MARL algorithms include IPPO\cite{zheng2020deep} and QTRAN\cite{son2019QTRAN}. For more details about the algorithm and hyper-parameter, see Appendix~\ref{OR algorithm} and Appendix~\ref{MARL algorithm}.

We present an overview of the experiment settings in Table~\ref{parameter in different challenge}. In our experiments, challenging tasks are derived from a standard task that follows general inventory management logic with realistic data settings. By modifying the setting, we develop versatile challenging tasks that primarily focus on scaling up, cooperation, competition, generalization and robustness. For more details on the environment design, refer to Appendix\ref{task detail design}.

\begin{table}
    \captionsetup{font=small}
    \caption{Experiments settings. "-" means the value is the same as the standard task, "stable" refers to stable contexts from real data, "add gap" implies constant variance with random mean changes, "add noise" keeps the mean constant but adds noise to increase variance, and "\#SKU * N" indicates N times the number of SKUs.}
    \label{parameter in different challenge}
    \resizebox{\textwidth}{!}{
        \begin{tabular}{cccccc}
            \toprule
            Task & \#Echelon & \#SKU & Capacity & Train contexts & Test contexts \\
            \midrule
            Standard & 1 & 200 & \#SKU * 100 & Stable & Stable \\
            \midrule
             Scaling up & - & $ 500,1000,2000 $ & - & - & - \\
            Cooperation & 2, 3 & - & - & - & -  \\
            Competition & - & - & \#SKU * 50, \#SKU * 25 & - & - \\
            Generalization & - & - & - & - & Add gap \\ 
            Robustness & - & - & - & Add noise & Add noise\\
            \bottomrule
        \end{tabular}
    }
\end{table}
We calculate the mean profit over all SKUs and warehouses, and then sum these mean profits over time steps to use as the evaluation metric. For the MARL algorithm, we select the best-performing model from the validation set and evaluate it on the test set. All training jobs are conducted using a single A100 graphics card. 

\subsection{Scaling up}
We present the results for scaling up experiments in Table~\ref{balance per SKU for scaling up}. The IPPO algorithm performs effectively when there is a small number of SKUs. However, its performance deteriorates as the number of agents increases, and it fails entirely when the number of SKUs reaches 2000. For QTRAN, although it yields good results, the training process demands significant time and GPU memory, making it resource-intensive.

\begin{table}[h]
    \captionsetup{font=small}
    \small
    \caption{Result of scaling up tasks.}
    \label{balance per SKU for scaling up}
    \centering
    \resizebox{\textwidth}{!}{
        \begin{tabular}{ccccccccccc}
            \toprule
            \multirow{2}{*}{Experiment} & \multicolumn{6}{c}{Mean profit} & \multicolumn{2}{c}{Time usage} & \multicolumn{2}{c}{Memory usage} \\
            & BS static & BS dynamic & ($s, S$) static & ($s, S$) hindsight & IPPO & QTRAN & IPPO & QTRAN & IPPO & QTRAN \\
            \midrule
            200 SKUs & 6.29k & 6.32k & 8.18k & 8.81k & 9.83k & 9.07k & 11.47h & 12.80h & 3.15G & 5.63G \\
            500 SKUs & 6.93k & 7.71k & 9.09k & 10.13k & 11.21k & 11.11k & 16.90h & 18.15h & 5.79G & 11.91G \\
            1000 SKUs & 5.64k & 6.37k & 7.92k & 8.85k & 8.2k & 9.44k & 23.60h & 34.58h & 10.39G & 22.61G \\
            2000 SKUs & 4.48k & 5.58k & 6.69k & 7.97k & 1.96k & 8.32k & 31.23h & 52.28h & 19.77G & 44.31G \\
            \bottomrule
        \end{tabular}
    }
\end{table}

\begin{figure}[h]
    \centering
    \includegraphics[width=\textwidth]{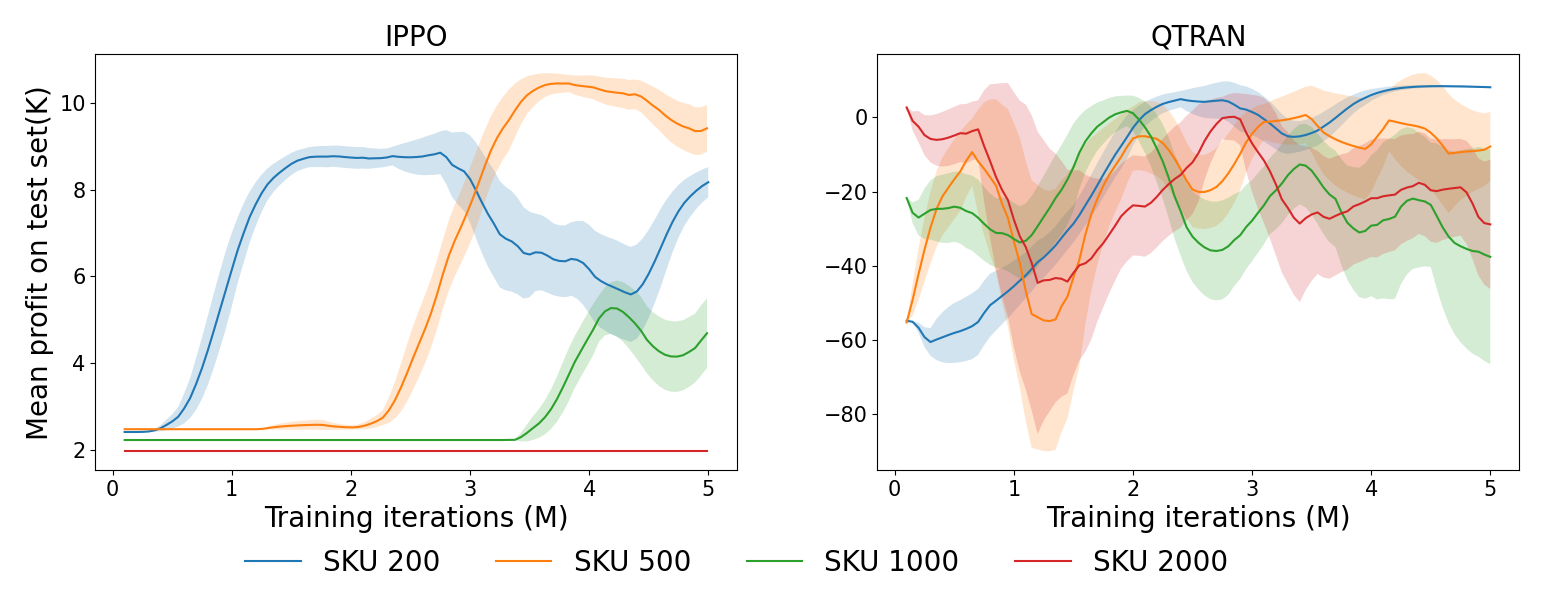}
    \caption{Mean profit in scaling up tasks.}
    \label{balance per SKU figure for scaling up}
\end{figure}

We examine the impact of scaling up on the MARL algorithm by displaying the profit on the test set during training iterations in Figure~\ref{balance per SKU figure for scaling up}. As the number of SKUs increases, finding an optimal strategy for IPPO becomes more challenging, resulting in convergence failure at 2,000 agents, even after 5 million iterations. Although QTRAN exhibits better performance, it encounters substantial training instability, which could present risks in real-world applications. This highlights the difficulties associated with increasing agent numbers in the training process.

\subsection{Competition}
We present the competition results in Table~\ref{balance per sku for different SKU count}. When capacity becomes lower, competition for capacity is more incentivized, leading to a greater impact on the base stock's static mode and the IPPO algorithm. As a result, these methods may face challenges in maintaining optimal performance under constrained capacity conditions.

\begin{table}[h]
    \captionsetup{font=small}
    \small
    \caption{Result of competition tasks.}
    \label{balance per sku for different SKU count}
    \centering
        \begin{tabular}{ccccccc}
            \toprule
            Experiment & BS static & BS dynamic & ($s, S$) static & ($s, S$) hindsight & IPPO & QTRAN \\
            \midrule
            Normal capacity & 6.29k & 6.32k & 8.18k & 8.81k & 9.83k & 9.07k \\
            Lower capacity & 5.01k & 3.64k & 6.13k & 7.06k & 6.46k & 6.74k \\
            Lowest capacity & -7.4k & 1.42k & 3.32k & 4.55k & -1.9k & 2.04k \\
            \bottomrule
        \end{tabular}
\end{table}

\begin{figure}[h]
    \centering
    \includegraphics[width=\linewidth]{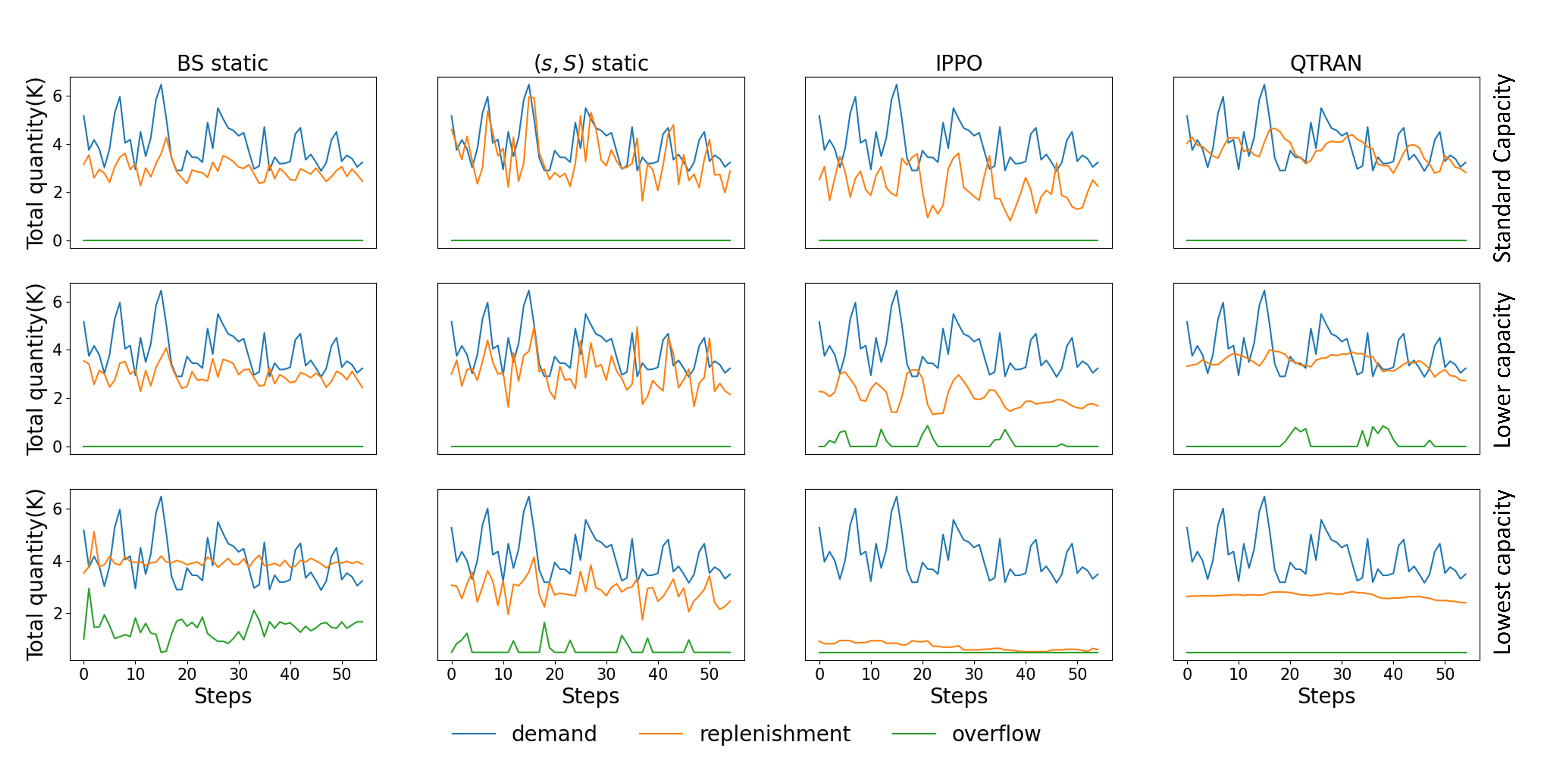}
    \caption{Policies for competition tasks. The demand, replenishment, and overflow represent the total quantity encompassing all SKUs.}
    \label{policy analysis for different capacity}
\end{figure}

We analyze algorithm policies by plotting them in Figure~\ref{policy analysis for different capacity}. As capacity decreases, BS strategy stays unchanged since it calculates stock levels per SKU without considering overall capacity, causing overflow and higher costs. IPPO reduces replenishment quantity, potentially avoiding short-term purchases to prevent losses but not maximizing long-term profit. Both ($s, S$) static and QTRAN algorithms show better stability in performance.

\subsection{Cooperation}
\begin{figure}[h]
    \centering
    \includegraphics[width=\linewidth]{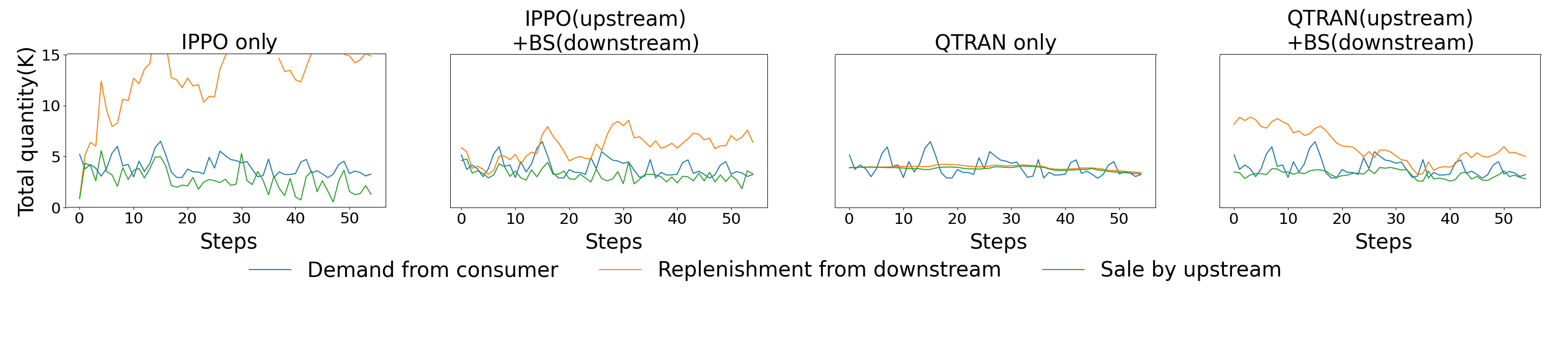}
    \caption{Policies for cooperation tasks. The demand, replenishment and sale represent the total quantity encompassing all SKUs.}
    \label{policy analysis for 2 echelons}
\end{figure}

We present the cooperation results in Table~\ref{balance per SKU for different echelons}. To assess the collaboration between upstream and downstream operations, we design two-echelon and three-echelon inventory management models. The MARL algorithms is not as effect as in single echelon, especially IPPO. To determine if learning both upstream and downstream strategies simultaneously results in poor performance, we introduce IPPO+BS and QTRAN+BS algorithms. These algorithms utilize the IPPO/QTRAN algorithm for the upstream warehouse and the BS algorithm for warehouses in other warehouse. The combined algorithm's performance significantly surpasses that of the pure OR and MARL algorithms. Based on these observations, we hypothesize that MARL performs well in a single warehouse, but encounters difficulties when managing both types of interactions.
\begin{table}[h]
    \captionsetup{font=small}
    \small
    \caption{Result of cooperation tasks.}
    \label{balance per SKU for different echelons}
    \centering
    \resizebox{\textwidth}{!}{
        \begin{tabular}{ccccccccc}
            \toprule
            Experiment & BS static & BS dynamic & ($s, S$) static & ($s, S$) hindsight & IPPO & IPPO + BS & QTRAN & QTRAN + BS\\
            \midrule
            Single echelon & 6.29k & 6.32k & 8.18k & 8.81k & 9.83k & - & 9.07k & - \\
            2 echelons & 9.15k & 7.76k & 9.1k & 9.68k & 6.72k & 9.86k & 9.88k & 10.54k \\
            3 echelons & 8.45k & 7.74k & 7.92k & 8.27k & 4.25k & 10.14k & 8.09k & 9.18k \\
            \bottomrule
        \end{tabular}
    }
\end{table}
We analyze algorithm policies by plotting them in Figure~\ref{policy analysis for 2 echelons}. Pure IPPO algorithm fulfill only a small demand portion. When the downstream warehouse strategy is fixed with BS, the upstream warehouse's demand fulfillment improves, enhancing overall performance. This indicates that insufficient information exchange between upstream and downstream entities results in less effective multi-layer strategy cooperation.

\subsection{Non-stationary context}
In MABIM, we take into account context factors such as demand, selling price, procurement cost, lead time, and more. In this experiment, we use demand as the context and design tests to evaluate the algorithm's capabilities for generalization and robustness. In generalization  experiments, we apply an offset to test set demand, creating a new pattern unseen in training. In robustness experiments, we add random noise to demand in the entire dataset, testing the model's ability to handle fluctuations.

As ($s, S$) hindsight algorithm directly optimizes on test data, hence is able to achieve the best performance. Its result will be used as the denominator to normalize results of other algorithms for better comparison. See Appendix~\ref{non-stationary task and result} for more details on non-stationary context tasks and results.

We display the relative performance of different algorithms in Figure~\ref{relative balance of ($s, S$) hindsight}. As the gap and noise levels increase, each algorithm is impacted to varying extents. IPPO outperforms other algorithms, whereas the static OR algorithm fares the worst, particularly in generalization tasks. This result illustrates that MABIM is capable of effectively measuring the generalization and robustness of a strategy.

\begin{figure}[h]
    \centering
    \includegraphics[scale=0.7]{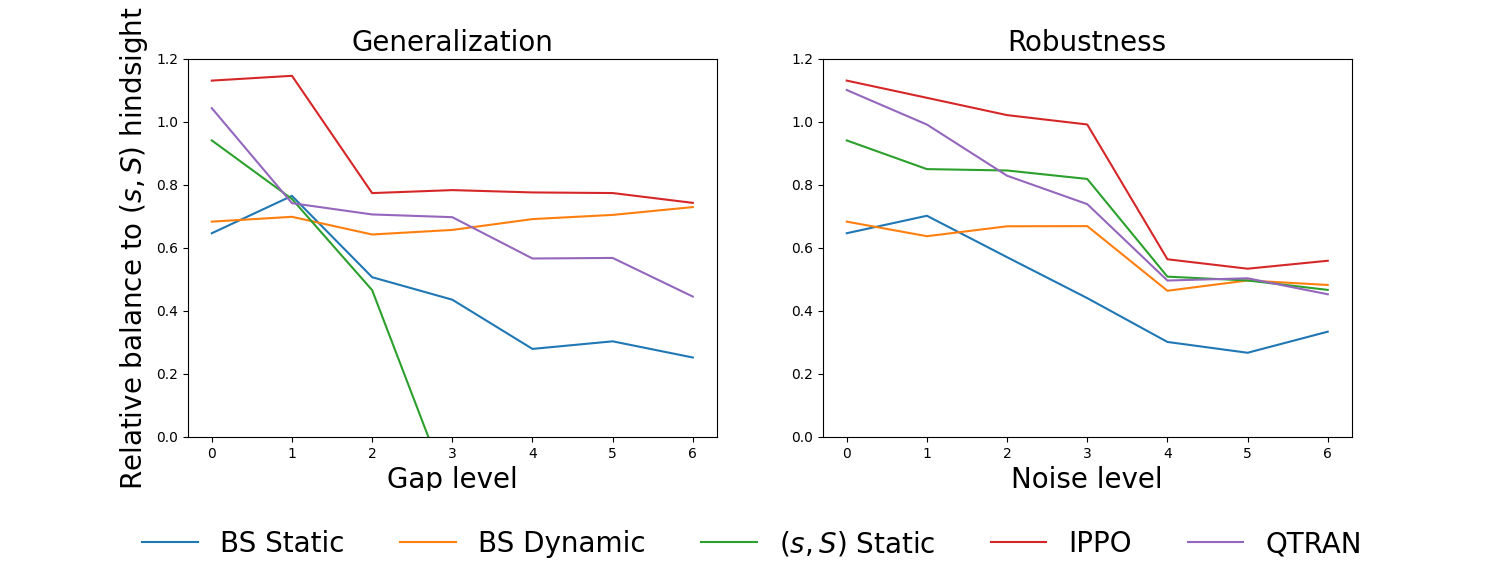}
    \caption{Relative profit of ($s, S$) hindsight.}
    \label{relative balance of ($s, S$) hindsight}
\end{figure}

\section{Conclusion}
In this paper, we introduce a MARL benchmark called MABIM, which can simulate a diverse range of challenging scenarios based on the inventory management problem. By utilizing MABIM, we develop various tasks that reveal limitations in existing MARL algorithms, such as unstable training with numerous agents, suboptimal performance under limited source, difficulties in cooperation with upstream and downstream, and the necessity for enhanced generalization and robustness. These findings highlight the importance of developing advanced MARL algorithms and demonstrate MABIM's potential as a valuable MARL benchmark. We believe that MABIM will significantly contribute to the progress of both inventory management and MARL research.


In future work, we aim to address the issues found in MABIM and expand its capabilities to better support MARL research. Potential extensions involve constructing tree-based or graph-based inventory management structures to evaluate agent communication abilities. Additionally, we will hide some SKU features to assess the MARL algorithm performance in partially observable settings.

\newpage
\bibliography{main}

\newpage
\appendix

\section{OR Algorithm}
\label{OR algorithm}
In our experiment, we use 2 classical algorithms, base stock policy and ($s, S$) policy.
\subsection{Base stock policy}
The base stock policy is a simple and efficient inventory management strategy, in which orders are placed to replenish the inventory when the stock quantity falls below the base stock level. This policy is considered a lower bound due to its simplicity and speed. The base stock level is calculated through programming, as demonstrated in Equation~\ref{base stock policy equation}.
\begin{equation*}
\label{base stock policy equation}
\begin{split}
{\bf max}\quad o_{i,j}^t &= \bar{p}_{i,j} \cdot S_{i,j}^t - \bar{c}_{i,j} \cdot S_{i+1,j}^t - \bar{h}_{i,j} \cdot I_{i,j}^{t+1} - \bar{c}_{i,j} \cdot T_{i,j}^0 - \bar{c}_{i,j} \cdot I_{i,j}^0 \\
{\bf s.t}\quad I_{i,j}^{t+1} &= I_{i,j}^t + S_{i+1,j}^{t-\bar{L}_{i,j}} - S_{i,j}^t \\
\quad T_{i,j}^{t+1} &= T_{i,j}^t - S_{i+1,j}^{t-\bar{L}_{i,j}} + S_{i+1,j}^t \\
S_{i,j}^t &= min(I_{i,j}^t, R_{i,j}^t) \\
T_{i,j}^0 &= \sum_{t=-\bar{L}_{i,j}}^{-1} S_{i+1,j}^t \\
z_{i,j} &= I_{i,j}^{t+1} + S_{i+1,j}^t + T_{i,j}^t \\
z_{i,j} &\in \mathbb{R^+}\\
\end{split}
\end{equation*}
In the equations mentioned above, $i$, $j$, and $t$ represent the indices of the warehouse, SKU, and time step, respectively. We also use the mean values $\bar{p}$, $\bar{c}$, $\bar{h}$, and $\bar{L}$ to denote the average selling price, procurement cost, holding cost, and lead time, respectively. The variables $S$, $R$, $I$, and $T$ represent the sale quantity, replenishment quantity, quantity in stock, and quantity in transit, respectively. Lastly, $o_{i,j}^t$ is the profit as the objective, and $z_{i,j}$ represents the base stock level, which is independently calculated for each product.

In the \textbf{static mode}, all base stock levels are calculated using historical data from the training set and consistently applied to the test set. These levels remain fixed throughout the test period. In the \textbf{dynamic mode}, the base stock levels are calculated using historical data and are updated at regular intervals. The levels are recalculated based on the updated historical data.

\subsection{($s, S$) Policy}
The ($s, S$) policy is an effective inventory management strategy wherein orders are placed to replenish inventory when the stock quantity reaches or falls below the base reorder level, denoted as $s$. The policy aims to restore the inventory to its maximum level, represented by $S$. Due to its efficacy, the ($s, S$) policy serves as a powerful baseline, and an optimal ($s, S$) pair is sought for each SKU in the dataset.
In \textbf{static mode}, ($s, S$) is searched in train set and in \textbf{hindsight mode}, ($s, S$) is searched in test set.

\section{MARL Algorithms}
\label{MARL algorithm}
In the training process of the MARL algorithm, some implementation details include:
\begin{itemize}
    \item The observation encompasses both SKU features and warehouse states. SKU features consist of quantity in stock, selling price, procurement cost, mean and standard deviation of historical demand, holding cost, order cost, and lead time. Warehouse states include total quantity in stock and remaining space, total profit in stock, total quantity in transit, and total profit in transit. All these features are normalized for optimal performance.
    \item During the training process, state and reward values are normalized using a rolling average and standard deviation. This approach enhances the model's learning and convergence capabilities.
    \item For replenishment actions, we employ multiples of the average demanded quantity from the past 21 days, which ensures that inventory levels are maintained based on recent demand trends, resulting in more effective inventory management decisions.
\end{itemize}
Table~\ref{hyperparameters for MARL} lists the training-related hyperparameters.
  
\begin{table}[h]
    \caption{Hyperparameters for MARL.}  
    \label{hyperparameters for MARL}  
    \centering  
    \begin{tabular}{lll}  
        \toprule  
        \textbf{Hyperparameter} & \textbf{IPPO} & \textbf{QTRAN} \\  
        \midrule  
        \#Epochs & 5020000 & 5020000 \\
        Discount rate & 0.985 & 0.985 \\  
        Optimizer & Adam & Adam \\  
        Optimizer alpha & 0.99 & 0.99 \\  
        Optimizer eps & 1e-5 & 1e-5 \\  
        Learning rate & 5e-4 & 5e-4 \\  
        Grad norm clip & 10 & 10 \\  
        Horizon & 21 & 21 \\
        Eps clip & 0.15 & - \\  
        Critic coef & 0.5 & - \\  
        Entropy coef & 0 & - \\  
        Accumulated episodes & 1 & 8 \\
        \bottomrule  
    \end{tabular}  
\end{table}  

\section{Details of Tasks Design}
\label{task detail design}
\paragraph{Standard task}
The design principles behind these environment parameters ensure that more reasonable purchasing strategies yield higher benefits. A reasonable strategy encompasses the following factors:  
\begin{itemize}  
    \item Most products have a purchase frequency of 2 to 10 steps.  
    \item There is no optimal strategy for products without replenishment.  
    \item In most cases, commodities are not purchased in excessive quantities at once (no more than 30 times the historical average daily sales volume).  
    \item The holding and storage costs account for 10\% to 25\% of the gross merchandise volume (GMV) for one year, where $GMV_{i,j} = \sum_{t=1}^{365}{s_{i,j}^t*p_{i,j}^t}$.  
\end{itemize}  

Based on ($s, S$) policy, we search the following hyperparameter as standard configuration in Table~\ref{standard configuration}. Rest feature including selling price, procurement cost, leading time are real dynamic data from external file.
\begin{table}[h]
    \caption{Standard configuration.}  
    \label{standard configuration}  
    \centering  
    \begin{tabular}{ll}  
        \toprule  
        Hyperparameter & value \\  
        \midrule
        \#SKU & 200 \\
        \#Warehouse & 1 \\
        Capacity & 20000 \\
        Order cost & 10 \\
        Holding cost & 0.002 + 0.001 \\
        Backlog cost & 0.1 * (selling price - procurement cost) \\
        Overflow cost & 0.5 * procurement cost \\
        \bottomrule
    \end{tabular}  
\end{table} 
\paragraph{Total tasks}
Based on the standard task, we build-in total of 51 tasks in MABIM. Seen in Table~\ref{total tasks}
\begin{table}[h]
    \caption{Total tasks. "$+$" indicates the extent to which the challenge is involved.}  
    \label{total tasks}  
    \centering  
    \resizebox{\textwidth}{!}{
        \begin{tabular}{lcccccc}  
            \toprule  
            Task name & Scaling up & Cooperation & Competition & Generalization & Robustness & More challenge\\
            \midrule
            sku50.single\_store.standard & & & & & & \\
            sku50.2\_stores.standard & & $+$ & & & & \\
            sku50.3\_stores.standard & & $+$ $+$ & & & & \\
            \midrule
            sku100.single\_store.standard & & & & & & \\
            sku100.2\_stores.standard & & $+$ & & & & \\
            sku100.3\_stores.standard & & $+$ $+$ & & & & \\
            \midrule
            sku200.single\_store.standard & & & & & & \\
            sku200.2\_stores.standard & & $+$ & & & \\
            sku200.3\_stores.standard & & $+$ $+$ & & & & \\
            \midrule
            sku500.single\_store.standard & & & & & & \\
            sku500.2\_stores.standard & $+$ & $+$ & & & & \\
            sku500.3\_stores.standard & $+$ & $+$ $+$ & & & & \\
            \midrule
            sku1000.single\_store.standard & & & & & & \\
            sku1000.2\_stores.standard & $+$ $+$ & $+$ & & & & \\
            sku1000.3\_stores.standard & $+$ $+$ & $+$ $+$ & & & & \\
            \midrule
            sku2000.single\_store.standard & & & & & & \\
            sku2000.2\_stores.standard & $+$ $+$ $+$ & $+$ & & & & \\
            sku2000.3\_stores.standard & $+$ $+$ $+$ & $+$ $+$ & & & & \\
            \midrule
            sku200.single\_store.lower\_capacity & & & $+$ & & \\
            sku200.single\_store.lowest\_capacity & & & $+$ $+$ & & \\
            sku200.2\_stores.lower\_capacity & & $+$ & $+$ & & \\
            sku200.2\_stores.lowest\_capacity & & $+$ & $+$ $+$ & & \\
            sku200.3\_stores.lower\_capacity  & & $+$ $+$ & $+$ & & \\
            sku200.3\_stores.lowest\_capacity  & & $+$ $+$ & $+$ $+$ & & \\
            \midrule
            sku200.single\_store.dynamic\_vlt & & & & & $+$ & \\
            sku200.2\_stores.dynamic\_vlt & & $+$ & & & $+$ & \\
            sku200.3\_stores.dynamic\_vlt & & $+$ $+$ & & & $+$ & \\
            \midrule
            sku200.single\_store.increase\_demand & & & & $+$ & $+$ & \\
            sku200.single\_store.decrease\_demand  & & & & $+$ & $+$ & \\
            \midrule
            sku200.single\_store.higher\_backlog & & & & & & higher constraints \\
            sku200.single\_store.highest\_backlog & & & & & & highest constraints \\
            \midrule
            sku200.single\_store.higher\_holding\_cost & & & $+$ & & &\\
            sku200.single\_store.highest\_holding\_cost & & & $+$ & & &\\
            \midrule
            sku200.single\_store.higher\_order\_cost & & & & & & lower action frequency \\
            sku200.single\_store.highest\_order\_cost & & & & & & lowest action frequency \\
            \midrule
            sku200.single\_store.low\_profit & & & & & & low action space \\
            sku200.single\_store.high\_profit & & & & & & high action space \\
            \midrule
            sku200.single\_store.higher\_overflow\_cost & & & & & & higher punishment \\
            sku200.single\_store.highest\_overflow\_cost & & & & & & highest punishment \\
            \midrule
            sku200.single\_store.add\_gap\_1 & & & & $+$ & & \\
            sku200.single\_store.add\_gap\_2 & & & & $+$ & & \\
            sku200.single\_store.add\_gap\_3 & & & & $+$ $+$ & & \\
            sku200.single\_store.add\_gap\_4 & & & & $+$ $+$ & & \\
            sku200.single\_store.add\_gap\_5 & & & & $+$ $+$ $+$ & & \\
            sku200.single\_store.add\_gap\_6 & & & & $+$ $+$ $+$ & & \\
            \midrule
            sku200.single\_store.add\_noise\_1 & & & & & $+$ & \\
            sku200.single\_store.add\_noise\_2 & & & & & $+$ & \\
            sku200.single\_store.add\_noise\_3 & & & & & $+$ $+$ & \\
            sku200.single\_store.add\_noise\_4 & & & & & $+$ $+$ & \\
            sku200.single\_store.add\_noise\_5 & & & & & $+$ $+$ $+$ & \\
            sku200.single\_store.add\_noise\_6 & & & & & $+$ $+$ $+$ & \\
            \bottomrule
        \end{tabular}  
    }
\end{table} 

\section{Visualization}
\label{visulization}
\begin{figure}[h]
    \centering
    \includegraphics[width=\textwidth]{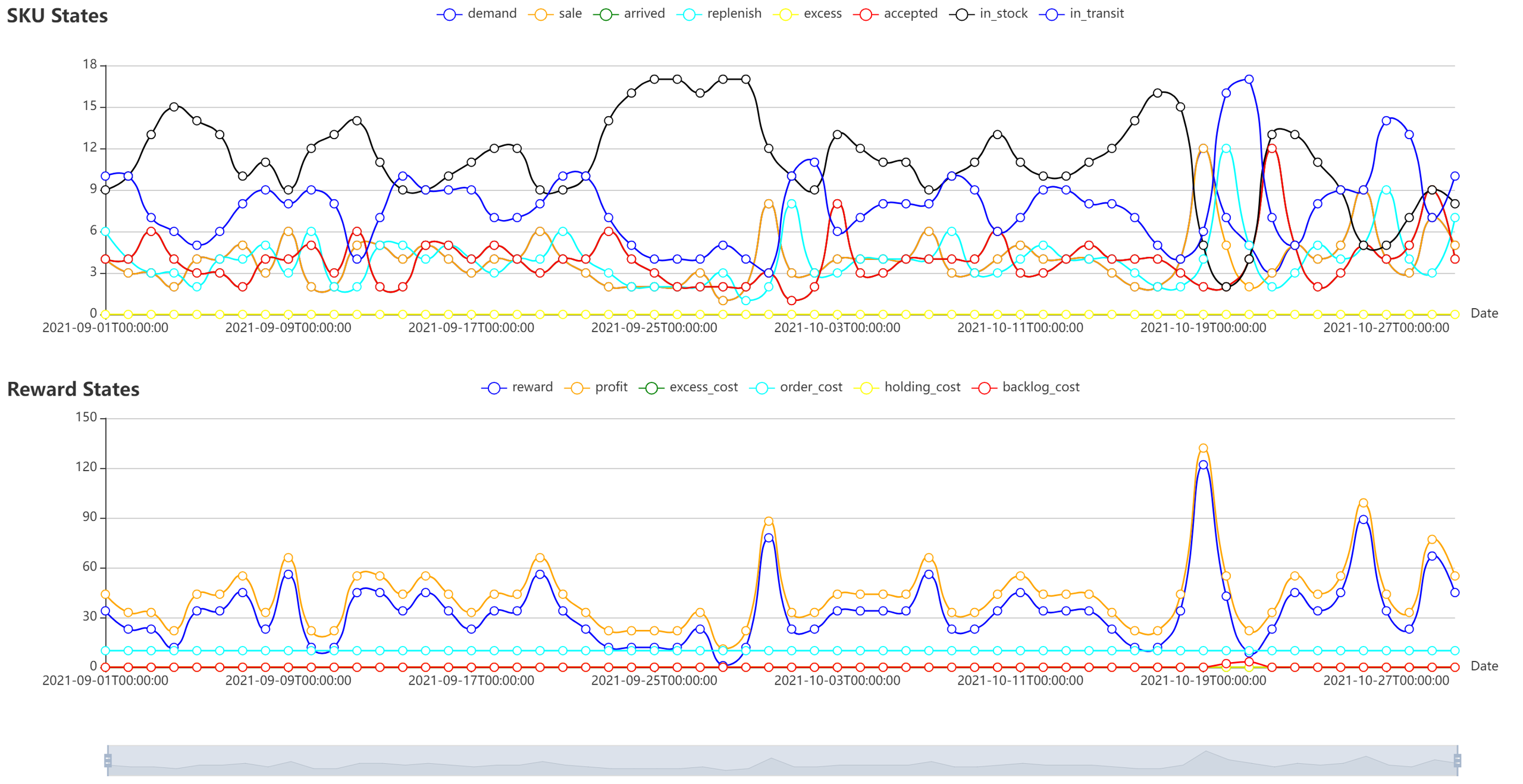}
    \caption{States of SKUs.}
    \label{states of SKU}
\end{figure}

MABIM displays the status of each product in every warehouse during each step, along with the overall warehouse status, in the form of a webpage, as shown in Figure~\ref{states of SKU}.
The information provided includes demand, sales, arrivals, acceptance, replenishment, excess, inventory, items in transit, and profit, as well as procurement costs, backlog costs, order costs, and holding costs.

\section{Non-stationary tasks and results}
\label{non-stationary task and result}
\begin{table}[h]
    \small
    \captionsetup{font=small}
    \caption{Result of non-stationary context tasks.}
    \label{balance per SKU for non-stationary context}
    \centering
    \resizebox{\textwidth}{!}{
    \begin{tabular}{cccccccc}
        \toprule
        Experiment & Gap level & BS static & BS dynamic & ($s, S$) static & ($s, S$) hindsight & IPPO & QTRAN \\
        \midrule
        \multirow{7}{*}{Generalization} & 0 (no gap) & 6.29k & 6.32k & 8.18k & 8.81k & 9.83k & 9.07k \\
        & 1 & 4.6k & 4.2k & 4.54k & 6.01k & 6.89k & 4.46k \\
        & 2 & 4.94k & 6.26k & 4.54k & 9.74k & 7.54k & 6.88k \\
        & 3 & 4.31k & 6.5k & -1.96k & 9.89k & 7.75k & 6.9k \\ 
        & 4 & 4.56k & 11.28k & -18.66k & 16.31k & 12.66k & 9.24k \\
        & 5 & 5.69k & 13.21k & -18.43k & 18.74k & 14.51k & 10.65k \\
        & 6 & 5.48k & 15.85k & -48.52k & 21.72k & 16.14k & 9.68k \\
        \midrule
        Experiment & Noise level & BS static & BS dynamic & ($s, S$) static & ($s, S$) hindsight & IPPO & QTRAN \\
        \midrule
        \multirow{7}{*}{Robustness} & 0(no noise) & 6.29k & 6.32k & 8.18k & 8.81k & 9.83K & 9.07K \\
        & 1 & 4.22K & 3.83K & 5.11K & 6.01K & 6.47K & 5.96K \\
        & 2 & 4.48K & 5.25K & 6.64K & 7.85K & 8.02K & 6.51K \\
        & 3 & 4.36K & 6.62K & 8.1K & 9.89K & 9.81K & 7.31K \\
        & 4 & 4.92K & 7.57K & 8.3K & 16.31K & 9.2K & 8.1K \\
        & 5 & 5.01K & 9.31K & 9.31K & 18.74K & 10.01K & 9.44K \\
        & 6 & 7.26K & 10.48K & 10.14K & 21.72K & 12.15K & 9.84k \\
        \midrule
    \end{tabular}
    }
\end{table}
\end{document}